\pdfoutput=1
\documentclass[twocolumn]{qartreport}
\usepackage{needspace}
\usepackage{supertabular}

\paperstyle{simple}
\papercolor{blue}

\title{QART: A Quantum-Classical Hybrid Architecture for Long-Horizon Reasoning -- Exploring a Conditional Path toward Quantum Scaling}

\author[1,2,3,*]{Lehao Lin}
\author[1,3,*]{Yuheng Cheng}
\author[1,2,3,*]{Guolong Liu}
\author[3]{Yao Li}
\author[3]{Xuning Tan}
\author[5]{Xiyuan Zhou}
\author[3]{Ruixi Zou}
\author[3]{Shi Wang}
\author[6]{Huan Zhao}
\author[5]{Wenxuan Liu}
\author[1,4]{Haifeng Wu}
\author[1,2,3,\dagger]{Junhua Zhao}

\affiliation[1]{QuantumMind}
\affiliation[2]{Shenzhen Institute of Artificial Intelligence and Robotics for Society}
\affiliation[3]{The Chinese University of Hong Kong, Shenzhen}
\affiliation[4]{Shenzhen Institute of Data Economy}
\affiliation[5]{Nanyang Technological University}
\affiliation[6]{The Hong Kong Polytechnic University}
\contribution[*]{These authors contributed equally and share first authorship.}
\contribution[\dagger]{The author is the corresponding author (zhaojunhua@cuhk.edu.cn).}

\newcommand{\reportlogos}{%
  \noindent\makebox[\linewidth][s]{%
    \includegraphics[width=\linewidth]{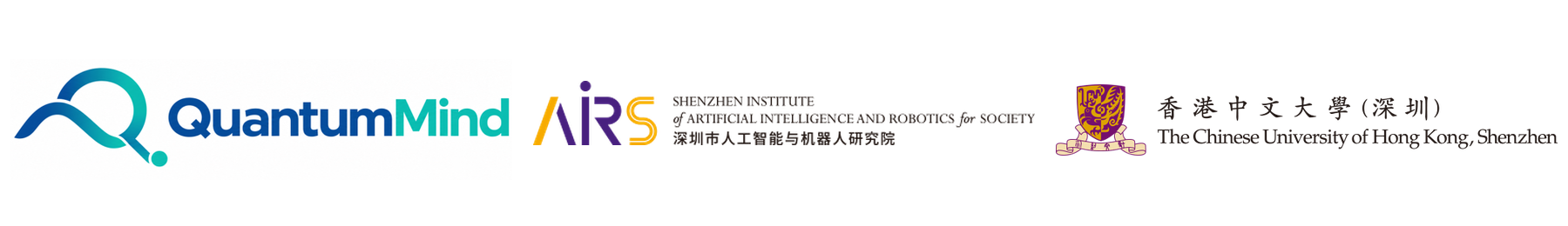}%
  }\par\vspace{0.02cm}%
}

\abstract{
Long-horizon tasks challenge large language models because early reasoning errors can alter subsequent decisions and compromise trajectory-level reliability.
We present QART, the Quantum-Augmented Reasoning Transformer, as a quantum--classical hybrid architecture combining a backbone language model with quantum encoding, CIM-based QUBO optimization, and quantum decoding as functional layers.
The architecture accommodates semantic information extracted from either hidden representations or model-generated text.
Detailed encoding and optimization procedures remain proprietary.

We establish that, under explicit assumptions, QART has an asymptotic advantage in optimal-path recovery over classical LLMs operating in a single-trajectory autoregressive setting.
For a common task family with aligned optimality and acceptance criteria, we show that the probability of retaining an acceptable autoregressive trajectory converges to zero when the cumulative conditional risk of irreversible reasoning errors diverges.
For QART, we combine a spectral ground-state certificate with semantic fidelity and explicit implementation events to derive a lower bound on task-optimal-path recovery. If the conditional probabilities associated with optimal-path coverage and semantic fidelity, spectral certification, dynamical reachability, and faithful readout remain uniformly positive as the reasoning horizon grows under a specified resource schedule, QART's recovery probability remains bounded away from zero. This establishes a conditional asymptotic reliability separation; the required uniform bounds are not implied by the architecture alone.

We report paired end-to-end measurements on $\tau^2$-Bench, $\tau^3$-Bench, SciCode, LHTB, DeepSWE, and Terminal-Bench 4.0 using DeepSeek V4 Flash, GLM-5.3, and GPT-5.5 xhigh in a Codex agent environment. Fourteen of fifteen evaluated backbone--benchmark pairs favor the QART hybrid configuration. Relative gains reach approximately 84.0\% on SciCode, 47.6\% on $\tau^3$-Bench, and 44.4\% on Terminal-Bench 4.0, while the DeepSeek V4 Flash-backbone configuration regresses by 7.8\% on DeepSWE.
These results provide initial system-level evidence; they do not directly validate the asymptotic separation.

We further formulate potential quantum scaling laws as conditional hypotheses linking effective optimization capacity to reliable reasoning horizons. Their quantum-advantage interpretation requires a demonstrated CIM quantum advantage over strong classical solvers and its transfer to end-to-end reasoning after all system overheads. 
}

\date{September 2026}

\newcommand{\QART}{QART\xspace}

\begin{document}
\maketitle
\raggedbottom

\section{Introduction}
\label{sec:introduction}

Transformer-based large language models \citep{vaswani2017attention} can write,
program, and answer complex questions, yet they remain brittle when a task
requires many dependent decisions
\citep{liu2023agentbench,valmeekam2022planbench,wang2024agentsurvey}. Each step
can introduce an error; once an early decision changes the state, assumptions,
or tool inputs seen by later steps, the deviation can propagate through the
remainder of the trajectory. Longer and more constrained tasks therefore place
substantial demands on consistency, task completion, and reliable use of
intermediate information.

\QART, the Quantum-Augmented Reasoning Transformer, addresses this setting as a
quantum--classical hybrid architecture. A backbone language model handles task
understanding, reasoning, and response generation, while an optimization
component processes task-relevant information and returns auxiliary information
for subsequent model activity. The QART functional layers comprise quantum
encoding, CIM-based QUBO optimization, and quantum decoding. Coherent Ising
machine technology provides the architecture's combinatorial optimization basis
when available \citep{yamamoto2017cim,mohseni2022isingmachines,mwamsojo2023optoelectronic}.
The report describes these functional roles without disclosing the proprietary
information-processing or optimization implementation.

The current report makes five contributions:
\begin{itemize}
  \item It specifies a modular QART architecture connecting a backbone language
        model with quantum encoding, CIM-based QUBO optimization, and quantum
        decoding functional layers.
  \item It formulates potential quantum scaling laws as conditional hypotheses
        based on solver-specific effective optimization capacity, and states the
        matched classical comparisons and resource-scale experiments required to
        test them.
  \item It derives a conditional asymptotic reliability separation: under
        uniform coverage, fidelity, spectral, and implementation assumptions,
        QART's optimal-path recovery stays positive while single-trajectory
        autoregressive success vanishes with horizon.
  \item It presents paired end-to-end measurements of classical baseline and
        QART hybrid configurations sharing the same backbone LLM on six
        long-horizon agent and coding benchmarks.
  \item It reports absolute score changes and relative gains for fifteen
        evaluated backbone--benchmark pairs, retaining the observed DeepSWE
        regression to delimit where the QART hybrid configuration is helpful.
\end{itemize}

The benchmark results provide initial system-level evidence for the evaluated
QART configurations, while variation across backbones and tasks shows that the
effect is not uniform. Optimization capacity is not interpreted as quantum
advantage: a potential quantum scaling law additionally requires sustained CIM
advantage over strong classical solvers and measurable transfer to end-to-end
reasoning after system overheads. These distinctions motivate controlled study
for domains with long operational, numerical, and policy constraints.

\section{Background and Positioning}
\label{sec:background}

\subsection{Language models and combinatorial optimization}

A language model represents information continuously and generates responses
incrementally. Complex tasks also involve discrete decisions subject to
multiple requirements, motivating the study of optimization-assisted
reasoning. Many discrete optimization problems admit Ising or quadratic
unconstrained binary optimization (QUBO) formulations
\citep{lucas2014ising,glover2019qubo}. These established formulations provide
the general background for the optimization technology considered here.

\subsection{Coherent Ising machines}

A coherent Ising machine represents Ising spins with the phases of coupled
nonlinear optical oscillators and evolves the network toward low-energy spin
configurations
\citep{mcmahon2016fully,inagaki2016coherent,yamamoto2017cim}.
QUBO and Ising formulations provide related mathematical descriptions of
binary optimization problems. Practical performance depends on problem
characteristics, hardware precision, noise, and system overhead
\citep{mohseni2022isingmachines,mwamsojo2023optoelectronic,bernalneira2025benchmarking}.
Consequently, the advertised oscillator or spin count alone is not a
sufficient measure of the problem size that can be solved reliably.

\subsection{From Neural Scaling Laws to Potential Quantum Scaling Laws}

Classical neural scaling laws describe how loss or task performance varies with
model size, training data, and computational resources
\citep{kaplan2020scaling,hoffmann2022training}. QART motivates a complementary
question: whether increasing the capacity for reliable optimization can extend
the reasoning horizon of a quantum--classical hybrid model under a specified
end-to-end resource budget.

An optimization-capacity--reasoning relationship, however, would not by itself
establish a quantum scaling law. Such a relationship could also arise when the
optimization layer is implemented with classical solvers. Interpreting it as a
scaling benefit enabled by quantum advantage would require evidence that CIM
technology provides a sustained advantage over strong classical optimization
alternatives on the relevant instance families, and that this advantage
translates into improved end-to-end reasoning performance.

We therefore use the term ``potential quantum scaling laws'' to denote
conditional research hypotheses. The present report establishes neither the
required CIM quantum advantage nor the proposed scaling relationships.
Section~\ref{sec:scaling} formulates these hypotheses and identifies the
conditions required for their validation.

\Needspace{6\baselineskip}
\section{QART Architecture}
\label{sec:architecture}

\subsection{System overview}

\begin{figure*}[t]
  \centering
  \includegraphics[width=\textwidth]{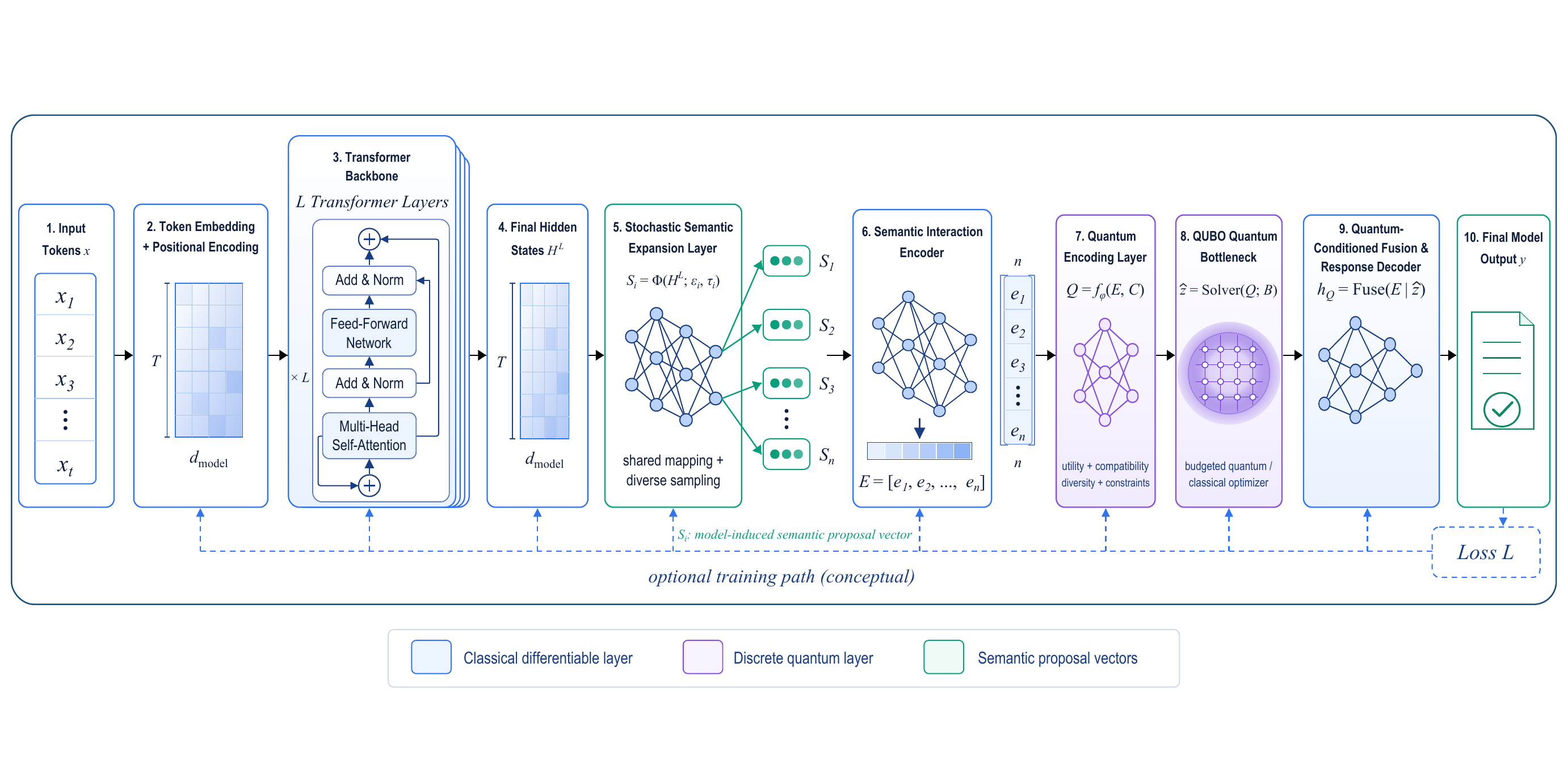}
  \caption{Architectural design of \QART, combining a Transformer language
  model with an optimization component in a quantum--classical hybrid
  architecture. The diagram is a conceptual hidden-representation route;
  an alternative model-generated-content interface $y_h$ is described in
  Section~\ref{sec:evaluation}. The solver returns a finite-budget candidate
  $\hat z=\operatorname{Solver}(Q;B)$, while $z^*\in\arg\min_z z^\top Qz$
  remains the theoretical reference optimum. The accompanying description
  focuses on functional roles rather than proprietary implementation details.}
  \label{fig:architecture}
\end{figure*}

Figure~\ref{fig:architecture} presents the \QART architecture. The language
model is responsible for understanding the task, reasoning, and producing the
response. An optimization module processes task-relevant information and
provides auxiliary information for subsequent model activity. The architecture
is designed to accommodate information from model representations or
model-generated content, without prescribing a single information source in
this functional description.
Consistent with the interface notation used in the theoretical analysis,
we write the resulting semantic information as $u_h$, obtained from either a
hidden representation $Z_h$ or generated text $y_h$.

\subsection{Optimization in the quantum--classical hybrid architecture}

The optimization component internally converts task-relevant information into
a form suitable for its selected optimization technology. The QART architecture
uses QUBO/Ising formulations and may use CIM technology as one technical basis
\citep{glover2019qubo,mohseni2022isingmachines}. The module transforms
optimization results into auxiliary information usable by subsequent
language-model processing. The purpose is to support more reliable task
execution and response generation.

\subsection{Functional roles and disclosure scope}

The functional division assigns language processing to the model and
optimization-related processing to an internal component of the QART
architecture. The model can use the returned information during subsequent
reasoning and generation. This organization describes the role of optimization
within the overall hybrid system.

The hidden-representation route in Figure~\ref{fig:architecture} is one
conceptual implementation of the interface boundary rather than a claim that
every deployment exposes model states.
Likewise, the dashed training path is a conceptual option and does not imply
that the benchmark runs trained or fine-tuned any of these modules.

The specific information acquisition, internal representations, optimization
problem construction, solution procedures, and result adaptation methods are
proprietary and are not detailed in this report. The evaluation therefore
focuses on observable end-to-end performance of the QART architecture.

\section{Potential Quantum Scaling Laws: A Conditional Framework}
\label{sec:scaling}

This section formulates potential quantum scaling laws for QART as conditional
hypotheses linking optimization resources to reliable reasoning horizons. The
proposed relationships are not established scaling laws, and the theoretical
results in Section~5 do not derive a universal power-law relationship between
CIM size and reasoning performance.

The proposed quantum-advantage interpretation requires three conditions. First,
CIM technology must demonstrate a sustained advantage over strong classical
solvers on optimization instances relevant to QART under a clearly specified
resource comparison. Second, the encoding, decoding, and subsequent model
execution must preserve enough of that advantage to improve end-to-end
reasoning performance after all associated overheads are included. Third,
controlled experiments across resource scales must establish a reproducible
relationship between effective optimization capacity and externally measured
reasoning performance.

The present report does not establish these conditions. The formulation below
identifies quantities and comparisons through which the potential scaling
behavior could be investigated.

\subsection{Resource Variables and Comparison Conditions}

Let $N_{\mathrm{CIM}}$ denote the available physical oscillator or spin
channels, $n$ the number of logical variables in an internal optimization
problem, and $B$ a specified end-to-end time budget. These quantities describe
resource scale without specifying the meaning or construction of the internal
variables. The end-to-end budget includes model execution, representation
processing, encoding, solver communication and execution, sampling, and
decoding.

Raw CIM spin count is not a standalone measure of useful optimization capacity.
Hardware precision, noise, connectivity, control procedures, and system
overhead can affect the problem sizes that are solved reliably
\citep{mohseni2022isingmachines,mwamsojo2023optoelectronic,bernalneira2025benchmarking}.
Increasing $N_{\mathrm{CIM}}$ therefore does not, by itself, imply improved
reasoning performance or an advantage over classical optimization.

Comparisons must specify the instance family, solution-quality target, success
criterion, time budget, and available hardware resources. Classical baselines
should include strong exact, heuristic, and spectral methods where applicable.
A practical CIM performance advantage and an advantage attributable to quantum
resources are distinct claims; evidence for the former alone is insufficient
to establish the latter.

\subsection{Effective Optimization Capacity}

Effective optimization capacity is defined for a specified solver and
evaluation protocol; it is not intrinsically a quantum quantity. Let $S$ denote
a solver configuration, and let $\mathcal I_n$ be a prescribed distribution
of optimization instances with $n$ logical variables. For a target objective
tolerance $\delta$, success probability $p_0$, and end-to-end budget $B$,
define
\begin{equation}
\begin{aligned}
 n_{\mathrm{eff}}^{S}(B,\delta,p_0)
 &= \sup\Bigl\{n:\Pr\!\bigl[\\
 &\quad F(\hat z_{S,B})-F(z^\star)\leq\delta
 \bigr]\geq p_0\Bigr\}.
\end{aligned}
\label{eq:effective-capacity}
\end{equation}
Here, $\hat z_{S,B}$ is the solution returned within the prescribed budget
under the evaluation protocol, and $z^\star$ is an optimal reference solution
for the same instance. Failure to return a valid solution within budget counts
as failure. The probability is evaluated over the specified instance
distribution and solver randomness. The objective scale, instance distribution,
and treatment of system overhead must be fixed when comparing solvers.

This quantity measures the largest reliably solvable problem size under the
stated conditions. It applies equally to CIM-based and classical optimization.
Comparing $n_{\mathrm{eff}}^{\mathrm{CIM}}$ with the capacities of strong
classical baselines can characterize an operational capacity advantage. Such an
advantage alone does not establish its quantum origin or its transfer to
reasoning performance.

\subsection{Conditional Scaling Hypotheses and Validation Requirements}

Let $h$ denote an externally defined reasoning horizon, such as the number of
required task stages specified independently of a model's generated trajectory.
For a fixed task family, backbone LLM, semantic interface, and evaluation
protocol, let $h_{\max}^{S}(B)$ be the largest horizon at which a configuration
using solver $S$ meets a specified task-performance threshold within budget
$B$.

A candidate relationship is
\begin{equation}
\begin{aligned}
 h_{\max}^{S}(B)
 &\approx C_S\left[n_{\mathrm{eff}}^{S}(B,\delta,p_0)\right]^{\beta_S},\\[-0.25ex]
 &\qquad C_S>0,\quad \beta_S>0.
\end{aligned}
\label{eq:horizon-scaling}
\end{equation}
while an externally measured task loss may admit a regime-specific fit of the
form
\begin{equation}
\begin{aligned}
 L_S(n_{\mathrm{eff}}^{S})
 &\approx L_{\infty,S}+A_S\left(n_{\mathrm{eff}}^{S}\right)^{-\alpha_S},\\[-0.25ex]
 &\qquad A_S>0,\quad \alpha_S>0.
\end{aligned}
\label{eq:loss-scaling}
\end{equation}

The parameters must be estimated empirically, and neither a power-law form nor
common parameters across solvers are assumed to hold universally. These
equations describe generic optimization-capacity scaling hypotheses. They could
apply to classical or CIM-based implementations and are not, by themselves,
evidence of quantum advantage.

A potential quantum scaling law would require a further connection: a
demonstrated CIM quantum advantage on the relevant instance families would need
to produce a sustained improvement in the end-to-end capacity--reasoning
relationship relative to strong classical implementations. That improvement
must remain observable after accounting for encoding, decoding, communication,
sampling, and model-execution costs. Solver-level advantage is therefore a
prerequisite for the proposed quantum-advantage interpretation, but it is not
sufficient to establish it.

The conditional reliability results in Section~5 identify requirements under
which optimal-path recovery can remain non-vanishing as the reasoning horizon
grows. They do not establish CIM quantum advantage, determine the exponents in
Eqs.~\eqref{eq:horizon-scaling}--\eqref{eq:loss-scaling}, or prove that
increasing physical spin count extends the reliable reasoning horizon.

The benchmark results in Section~7 compare model configurations and task
families. They do not constitute a controlled scaling study or a comparison
against matched classical optimization layers, and they are not used to fit
Eqs.~\eqref{eq:horizon-scaling}--\eqref{eq:loss-scaling}. Validation requires
experiments across resource scales that jointly measure optimization capacity,
end-to-end cost, and reasoning performance for CIM-based and strong classical
implementations. Failure to observe a sustained CIM advantage, or failure of
that advantage to translate into improved reasoning performance, would
undermine the proposed quantum-advantage interpretation.

In these equations, $S$ denotes a specific solver configuration, while
$C_S,\beta_S,L_{\infty,S},A_S,\alpha_S$ are parameters to be estimated.
Solver-specific parameters avoid assuming without validation that classical
methods and CIM have identical scaling curves.

\section{Theoretical Foundations of Global Reasoning Reliability}
\label{sec:theory}

This section develops a conditional guarantee for recovering globally optimal reasoning paths in QART. Under a diverging cumulative risk of irreversible reasoning errors, the probability that a single autoregressive trajectory remains globally correct converges to zero. A spectral ground-state certificate for coherent Ising machines, combined with semantic fidelity, dynamical reachability, and faithful readout, yields a positive lower bound for QART whenever the associated conditional probabilities remain uniformly positive with reasoning depth. The analysis accommodates semantic information obtained from hidden representations or generated text. It specifies mathematical interface properties without prescribing proprietary encoding or optimization procedures. Complete proofs, robustness conditions, and a notation table are provided in Appendix~A.

The theoretical question is whether replacing successive local commitments with a globally optimized semantic decision can prevent the probability of selecting a correct reasoning path from vanishing as tasks become longer. We analyze a family of tasks indexed by reasoning horizon $h$. The horizon counts required reasoning structure rather than input or output tokens. It is distinct from the encoded problem dimension $n_h$ and from the physical execution time.

The argument has three components. First, the chain rule gives a precise error condition under which a single autoregressive path fails asymptotically. Second, an existing Nature Communications result supplies a sufficient certificate for recovering an Ising ground state. Third, semantic fidelity transfers that certificate to reasoning-path optimality, while explicit conditional probabilities account for representation and physical implementation. The resulting guarantee is conditional on these probabilities remaining uniformly bounded away from zero as the reasoning horizon grows under a specified resource schedule; establishing these bounds remains part of the research problem.

\subsection{Vanishing Correct-Path Probability in Single-Trajectory Autoregressive Reasoning}
\label{sec:ar-reliability}

For each horizon $h$, consider solvable tasks $q_h$ with $m(h)$ critical decisions and a nonempty set $\mathcal A_h$ of acceptable complete reasoning trajectories. A trajectory contains the decisions relevant to its semantic validity; multiple equivalent correct solutions may belong to $\mathcal A_h$. The analysis does not require a particular wording or token sequence. An autoregressive policy commits to successive decisions conditioned on the preceding history.

Let $C_{h,j}$ be the event that, after critical decision $j$, the committed prefix can still be extended to an element of $\mathcal A_h$ under the allowed continuation rules. Set $C_{h,0}=\Omega$. For an irreversible single-trajectory process these events are nested. Define
\begin{equation}
 e_{h,j}=1-\Pr(C_{h,j}\mid C_{h,j-1}).
 \label{eq:theory-risk}
\end{equation}
The probability is over the specified task distribution and policy randomness. Thus the risks are conditional on survival, and may include dependence between errors. The final critical decision includes completion, so that $C_{h,m(h)}$ is exactly the event of producing an acceptable complete trajectory.

\textbf{Proposition 1 (Vanishing correct-path probability).} Suppose $m(h)\to\infty$ and the cumulative conditional exit risk $E_h$ diverges, where
\begin{equation}
 E_h=\sum_{j=1}^{m(h)}e_{h,j}.
 \label{eq:cumulative-risk}
\end{equation}
Then the single-trajectory success probability satisfies
\begin{equation}
 P_{\rm AR}^{\rm path}(h)=\prod_{j=1}^{m(h)}(1-e_{h,j}),
 \label{eq:ar-product}
\end{equation}
and
\begin{equation}
 P_{\rm AR}^{\rm path}(h)\le e^{-E_h}\longrightarrow0.
 \label{eq:ar-bound}
\end{equation}
The proof uses only the chain rule and $\log(1-x)\le -x$; see Appendix~\ref{app:prop-proof}. A uniform critical-step risk $e_{h,j}\ge\varepsilon>0$ gives the familiar bound $P_{\rm AR}^{\rm path}(h)\le(1-\varepsilon)^{m(h)}$. Independent errors are not assumed. Prior work on compositional tasks reports related degradation with increasing task complexity~\citep{dziri2023faith}; the proposition here states the exact condition used in this chapter.

The result is about loss of an acceptable trajectory, not every locally imperfect token. A recoverable mistake is not an exit event. Effective backtracking, global verification, or sufficiently decreasing conditional risk may invalidate the divergence assumption. In particular, the classical-baseline condition in the Section 6 uses an agent environment, and is not automatically an instance of this restricted autoregressive baseline.

\subsection{A Spectral Sufficient Condition for Ising Ground-State Recovery}
\label{sec:spectral-certificate}

Wang et al.~\citep{wang2023bifurcation} established a sufficient synchronization condition under which the first bifurcation of a coherent Ising machine identifies an Ising ground state. We use its spectral form and state additional nondegeneracy assumptions to avoid ambiguity in the first mode and in sign readout.

Let $G_h$ be a nonzero, real, symmetric, zero-diagonal matrix of size $n_h\times n_h$. Define
\begin{equation}
 H_h(\boldsymbol\sigma)=-\frac12\boldsymbol\sigma^{\mathsf T}G_h\boldsymbol\sigma,
 \qquad \boldsymbol\sigma\in\{-1,+1\}^{n_h}.
 \label{eq:ising-theory}
\end{equation}
Let $H_{0,h}$ be its minimum energy and $H_{1,h}$ the smallest distinct energy above it. Put $\Delta H_h=H_{1,h}-H_{0,h}>0$. Ground states may be degenerate. The gap is to the lowest non-ground-state energy, not between two arbitrarily ordered configurations.

The normalized deterministic dynamics considered in~\citep{wang2023bifurcation} has the form
\begin{equation}
 \dot{\mathbf x}=(p-1)\mathbf x-\mathbf x^{\circ3}+\xi G_h\mathbf x,
 \label{eq:cim-dynamics}
\end{equation}
with $\xi>0$, pump parameter $p$, and componentwise cubic nonlinearity. The origin loses stability at
\begin{equation}
 p_{0,h}=1-\xi\lambda_{\max,h}.
 \label{eq:cim-threshold}
\end{equation}
Assume the largest eigenvalue is simple. Let $\mathbf v_h$ be a corresponding unit eigenvector with no zero coordinates, and let $\boldsymbol\sigma_h^c=\operatorname{sign}(\mathbf v_h)$. Define
\begin{equation}
 \alpha_h^2=\frac{(\mathbf v_h^{\mathsf T}\boldsymbol\sigma_h^c)^2}{n_h},
 \qquad w_h=\lambda_{\max,h}-\lambda_{\min,h}>0.
 \label{eq:nc-overlap}
\end{equation}

\textbf{Theorem NC (Spectral ground-state certificate; adapted from~\citep{wang2023bifurcation}).} Under the preceding conditions, if
\begin{equation}
 \alpha_h^2>1-\frac{2\Delta H_h}{n_hw_h},
 \label{eq:nc-certificate}
\end{equation}
then $\boldsymbol\sigma_h^c$ and $-\boldsymbol\sigma_h^c$ attain the ground-state energy. Appendix~\ref{app:spectral-proof} provides a self-contained spectral proof of the certificate used here.

At a simple first bifurcation, the normalized emerging branch approaches $\pm\mathbf v_h$. At a finite pump increment, the nonlinear equilibrium need not be exactly proportional to that eigenvector. Consequently, a physical application requires the observed branch to retain the certified signs. Appendix~\ref{app:spectral-proof} makes this distinction explicit. Moreover, an exactly zero initial state stays zero in the deterministic equation; nonzero fluctuations or perturbations must initiate departure from the origin. The spectral certificate alone gives no physical arrival probability or time guarantee.

\subsection{Semantic Interface and the Definition of Task Optimality}
\label{sec:semantic-interface}

The semantic interface can operate with accessible hidden representations $Z_h$ or with generated text $y_h$. We denote the resulting semantic information by $u_h$ and write the two interfaces separately:
\begin{equation}
 u_h=\Phi_{\rm hid}(Z_h,q_h),
 \label{eq:semantic-interface-hidden}
\end{equation}
\begin{equation}
 u_h=\Phi_{\rm text}(y_h,q_h).
 \label{eq:semantic-interface-text}
\end{equation}
These alternatives do not assert that every evaluated system exposes hidden states. The representation space $\mathcal U_h$ may be a vector space or an abstract measurable space. No particular extraction layer, feature dimension, or construction procedure is needed in the proof.

An abstract encoder and decoder satisfy
\begin{equation}
 G_h=\mathcal E(u_h,q_h),
 \label{eq:semantic-encoder}
\end{equation}
\begin{equation}
 D_h:\{-1,+1\}^{n_h}\longrightarrow\Pi_h\cup\{\bot\}.
 \label{eq:semantic-decoder}
\end{equation}
Here $\Pi_h$ is the declared finite admissible path space, and $\bot$ indicates invalid decoding. The task-level objective $L_h:\Pi_h\to\mathbb R$ is specified independently of the encoded energy, with
\begin{equation}
 \Pi_h^\star=\underset{\pi\in\Pi_h}{\arg\min}\,L_h(\pi).
 \label{eq:optimal-path-set}
\end{equation}
For each realized encoding, at least one element of $\Pi_h^\star$ must have a valid spin representation. This coverage requirement is separate from finding the lowest encoded energy. If $\Pi_h$ is a restricted space, the theorem establishes optimality within that space. Extending the claim to all admissible reasoning trajectories requires coverage of a genuine unrestricted optimum.

The zero-field Ising form is a theoretical representation; linear fields can be handled by a standard reference-spin reformulation~\citep{lucas2014ising}. Any resulting increase in dimension is included in $n_h$. The decoder must resolve the global sign symmetry consistently, for example through relative signs to the reference spin. No coefficient recipe or internal representation structure is specified. For a restricted hardware or encoding class, applicability of this representation must be established explicitly.

\subsection{Semantic Fidelity and the Transfer from Ground States to Optimal Reasoning Paths}
\label{sec:semantic-fidelity}

An energy minimum is useful only if energy ordering preserves the relevant task objective. Let $L_h^\star$ be the minimum of $L_h$. Assuming at least one non-optimal admissible path, define the task-objective gap
\begin{equation}
 \gamma_h=\min_{\pi\notin\Pi_h^\star}\bigl[L_h(\pi)-L_h^\star\bigr]>0.
 \label{eq:objective-gap}
\end{equation}
where the minimum is over $\Pi_h$. If every admissible path is optimal, the ordering requirement below is vacuous, while validity remains necessary.

\textbf{Assumption A (Coverage and semantic fidelity).} At least one optimal path has a valid encoding. For some $a_h>0$, $b_h\in\mathbb R$, and $\epsilon_h\ge0$, every valid spin state satisfies
\begin{equation}
 \left|H_h(\boldsymbol\sigma)-a_hL_h(D_h(\boldsymbol\sigma))-b_h\right|\le\epsilon_h,
 \qquad a_h\gamma_h>2\epsilon_h.
 \label{eq:semantic-fidelity}
\end{equation}
Every invalid state has energy strictly above the minimum valid energy. These requirements apply to the full represented state space, not merely a few sampled states.

\textbf{Lemma 1 (Ground-state-to-optimal-path transfer).} Under Assumption A, every ground state of $H_h$ decodes to an element of $\Pi_h^\star$.

To see the mechanism, compare an optimal encoding with a valid non-optimal encoding. Their energy difference is at least $a_h\gamma_h-2\epsilon_h>0$. Invalid configurations cannot be minima by assumption. Appendix~\ref{app:semantic-proof} gives the complete proof and an extension to near-optimal energy solutions.

Assumption A permits imperfect encoding while protecting the optimum. It is a strong, substantive property: high correlation between sampled path scores and energies does not prove a uniform bound. Defining the task objective retrospectively from the Ising energy would make the claim semantically empty. The objective, feasibility criteria, and reference judgments therefore require independent specification. The proof establishes ground-state inclusion in the optimal-path set; it does not require every optimal path to have exactly the same encoded energy.

\subsection{A Lower Bound on QART Global-Optimal-Path Recovery}
\label{sec:qart-bound}

Fix a task distribution at horizon $h$, the semantic interface, an encoding procedure, a physical configuration, and an end-to-end budget $B(h)$. Probabilities include the randomness of task sampling, semantic generation, encoding when stochastic, initial fluctuations, dynamical noise, and readout. For a fixed task and deterministic encoding, an encoding condition is deterministic; randomness must not be assigned to it without an explicit source.

Define the following events. $\mathsf M_h$ means that Assumption A holds. Conditional on $\mathsf M_h$, $\mathsf S_h$ means that the encoded Ising coupling matrix satisfies all hypotheses of Theorem NC. Conditional on both, $\mathsf B_h$ means that the implemented dynamics reaches a state with one of the certified sign patterns by the measurement deadline and retains those signs until readout. Finally, $\mathsf R_h$ means that measurement and decoding preserve that certified solution. A precise sign-neighborhood definition of $\mathsf B_h$ is given in Appendix~\ref{app:theorem-proof}.

For compactness, define the conditional factors
\begin{align}
 r_M(h)&=\Pr(\mathsf M_h), \label{eq:factor-rM}\\
 r_S(h)&=\Pr(\mathsf S_h\mid\mathsf M_h), \label{eq:factor-rS}\\
 r_B(h)&=\Pr(\mathsf B_h\mid\mathsf M_h,\mathsf S_h), \label{eq:factor-rB}\\
 r_R(h)&=\Pr(\mathsf R_h\mid\mathsf M_h,\mathsf S_h,\mathsf B_h). \label{eq:factor-rR}
\end{align}

\textbf{Theorem 1 (Conditional QART recovery bound).} Let $P_{\rm QART}^{\rm opt}(h)$ be the probability that the decoded output belongs to $\Pi_h^\star$. Then
\begin{equation}
 P_{\rm QART}^{\rm opt}(h)\ge r_M(h)r_S(h)r_B(h)r_R(h).
 \label{eq:qart-bound}
\end{equation}
If, for all sufficiently large $h$, each factor is bounded below by a positive constant $\rho_i$ independent of $h$, then
\begin{equation}
 p_\star=\rho_M\rho_S\rho_B\rho_R>0,
 \label{eq:uniform-product}
\end{equation}
\begin{equation}
 \liminf_{h\to\infty}P_{\rm QART}^{\rm opt}(h)\ge p_\star.
 \label{eq:uniform-bound}
\end{equation}
Theorem NC and Lemma 1 imply success on the intersection of the four events. The chain rule then yields Eq.~\eqref{eq:qart-bound}, without assuming independence. Appendix~\ref{app:theorem-proof} gives the proof. Positive constants are sufficient, not necessary: the system may also succeed on instances that lack the NC certificate, so a small certified lower bound is not a prediction of low total success.

The theorem exposes the obligations required for an asymptotic guarantee. It does not derive their uniformity from the QART architecture alone. In particular, conditional fidelity of a representation cannot compensate for a coverage probability that vanishes with horizon.

\subsection{Conditional Reliability Separation and Its Interpretation}
\label{sec:reliability-separation}

To compare correct reasoning probabilities, the two systems must address the same task family and acceptance criterion. Suppose every objective minimizer in $\Pi_h^\star$ is an acceptable path in $\mathcal A_h$, and the required coverage and implementation assumptions hold uniformly. Combining Proposition~1 and Theorem~1 yields
\begin{equation}
 P_{\rm AR}^{\rm path}(h)\longrightarrow0,
 \label{eq:separation-ar}
\end{equation}
\begin{equation}
 \liminf_{h\to\infty}P_{\rm QART}^{\rm path}(h)\ge p_\star.
 \label{eq:separation-qart}
\end{equation}
Consequently, the limiting inferior of the difference between the two path-success probabilities is at least $p_\star$. This is a conditional reliability separation between sequential commitment and certified global selection. It requires both the autoregressive cumulative-risk condition and QART's uniform conditional bounds.

A selected path is not an executed answer. Let $\mathsf X_h$ mean that subsequent tool use, numerical computation, and response generation produce a correct final outcome. A sufficient extension is a uniform bound $\rho_X>0$ on its probability conditional on the four-event certified success intersection. For all sufficiently large $h$, the path bound then gives
\begin{equation}
 P_{\rm QART}^{\rm answer}(h)\ge p_\star\rho_X.
 \label{eq:answer-bound}
\end{equation}
Without this extra condition, Eq.~\eqref{eq:uniform-bound} concerns path recovery only. Likewise, a proxy objective can be optimized perfectly while failing the independent task verifier. Objective alignment must therefore be stated separately from optimization accuracy.

These results neither establish quantum computational advantage nor show that general QUBO instances are efficiently solvable. The NC spectral certificate can also support classical spectral recovery on the certified instances. Its value here is a bridge from a mathematically specified optimization property to semantic reliability. Any claim of hardware advantage requires independent comparisons against strong classical alternatives, including spectral methods when applicable, under matched end-to-end budgets.

\subsection{Robustness and Resource Conditions for a Non-Vanishing Bound}
\label{sec:robustness-resources}

Let the implemented matrix be $\widetilde G_h=G_h+\Delta G_h$, with a symmetric perturbation and spectral norm $\eta_h=\|\Delta G_h\|_2$. For any binary spin state,
\begin{equation}
 \left|H_{\widetilde G_h}(\boldsymbol\sigma)-H_{G_h}(\boldsymbol\sigma)\right|\le\frac{n_h}{2}\eta_h.
 \label{eq:perturbation-energy}
\end{equation}
If $n_h\eta_h<\Delta H_h$, every perturbed ground state belongs to the original ground-state set. Degeneracy within that set may split, so equality of the two ground-state sets is not guaranteed. A separate semantic condition can preserve task-optimal decoding even when the exact ground-state identity changes; Appendix~\ref{app:perturbation} states both results.

For the certified sign pattern, useful quantities include the eigenvalue separation $\kappa_h=\lambda_{\max,h}-\lambda_{2,h}$, the smallest eigenvector magnitude $\mu_h=\min_i|v_{h,i}|$, and the strict certificate slack
\begin{equation}
 c_h=\Delta H_h-\frac{n_hw_h}{2}(1-\alpha_h^2)>0.
 \label{eq:certificate-slack}
\end{equation}
Positive finite-instance margins provide local robustness, but do not imply that tolerable perturbations remain constant as the problem grows. Indeed $\mu_h\le n_h^{-1/2}$ for a unit vector, so demanding a positive dimension-independent lower bound on $\mu_h$ would be impossible when $n_h\to\infty$. The relevant requirement is a controlled ratio between perturbation magnitude and the shrinking margin, or an appropriate physical amplitude scale.

The budget $B(h)$ must include representation, encoding, communication, physical evolution, sampling, and decoding. If those resources are fixed while $n_h$ and task complexity grow without bound, the uniform probabilities cannot simply be presumed. The hardware-aware scaling analysis should therefore study which resource schedules preserve them. Equations~\eqref{eq:qart-bound}--\eqref{eq:uniform-bound} do not establish a universal power law between oscillator count and reasoning horizon.

\section{Evaluation Setup}
\label{sec:evaluation}

\subsection{Paired Comparison of Classical and QART Hybrid Models}

We compare two model configurations built around the same backbone LLM: a
classical baseline and a QART quantum--classical hybrid model. The classical
baseline uses the backbone LLM in the Codex agent environment. The QART hybrid
model incorporates quantum encoding, CIM-based QUBO optimization, and quantum
decoding as functional layers within its reasoning architecture, with the same
LLM serving as its classical backbone.

In the QART configuration, the quantum encoding layer maps task-relevant
information into an optimization representation; the CIM-based QUBO layer
searches for low-objective-value configurations; and the quantum decoding layer
transforms the returned solution into information used in subsequent model
reasoning and generation. Here, QART denotes the complete hybrid architecture,
while these three layers constitute its quantum-assisted reasoning component.
Their proprietary implementations are not disclosed in this report.

\subsection{Information interface and training status}

The conceptual architecture in Figure~\ref{fig:architecture} supports two
functional ways to provide task-relevant information to the optimization
component: a representation interface that can expose model states, or a text
interface that consumes model-generated content. The language models and the QART
architecture were used at inference time under the fixed evaluation protocol;
no fine-tuning or policy-gradient update was performed during evaluation.
This implementation scope is narrower than the conceptual set of interfaces
shown in Figure~\ref{fig:architecture} and is the basis for the end-to-end
comparison below.

The CIM hardware used in the QART evaluation was provided by QBoson and had
1,000 computational qubits.

We evaluate DeepSeek V4 Flash and GLM-5.3 at maximum reasoning effort and
GPT-5.5 at xhigh effort. Table~\ref{tab:results} reports the measured scores
for each evaluated backbone--benchmark pair.

The reported values are \emph{our measured scores}, produced in the Codex agent
environment with the harness and protocol described here. They are not model
vendors' official scores, and this report does not mix public leaderboard
numbers into the paired comparison. Differences from external scores can arise
from the agent implementation, harness, prompt and tool environment, available
background material, retry policy, and aggregation method.

Each paired comparison uses the same evaluated backbone LLM and benchmark
verifier within the Codex agent environment. The results therefore compare the
original classical model configuration with a QART hybrid realization based on
that backbone. Throughout Sections~6 and~7, these configurations are labeled
``Classical baseline'' and ``QART hybrid'', respectively. All reported scores
are our own end-to-end measurements under the evaluation setup described here;
they are not model-vendor scores or imported public leaderboard results.

\subsection{Benchmarks}

The six benchmarks exercise different forms of long-range dependency:
\begin{itemize}
  \item $\tau^2$-Bench covers multi-turn Airline, Retail, and Telecom tasks in
        which an agent and user jointly update a shared environment
        \citep{yao2024taubench,barres2025tau2}.
  \item $\tau^3$-Bench uses 97 Banking tasks that combine unstructured policy
        retrieval with multi-step account, dispute, and permission operations
        \citep{artificialanalysis2026tau3}.
  \item SciCode contains scientist-authored research programming problems. It contains 291 subproblems; three with provided code substeps
        are excluded, so reported success rates use 288 as the denominator.
        Our setting evaluates the remaining subproblems without background
        context, one run per subproblem \citep{tian2024scicode}.
  \item LHTB contains 46 long-running terminal tasks and uses dense reward. Each
        condition is the mean of three clean per-task runs \citep{li2026lhtb}.
  \item DeepSWE contains 113 long-horizon software-engineering tasks evaluated
        by handwritten behavioral verifiers \citep{huang2026deepswe}.
  \item Terminal-Bench~4.0 is evaluated on 63 CPU tasks through Harbor with its
        built-in Codex agent and maximum thinking effort
        \citep{merrill2026terminalbench,terminalbench2026v4}. We exclude tasks
        requiring GPU hardware or GPU-specific software, retaining the CPU-only
        subset.
\end{itemize}

\subsection{Metrics and reporting}

Each benchmark's verifier supplies the primary score, with higher values
indicating better performance. Let $s_{\mathrm{base}}$ denote the score of the
classical baseline and $s_{\mathrm{QART}}$ the score of the QART hybrid model
constructed with the same backbone. The absolute score change is
\[
\Delta=s_{\mathrm{QART}}-s_{\mathrm{base}},
\]
and the relative gain over the classical baseline is
\begin{equation}
  g_{\mathrm{rel}}
  = \frac{s_{\mathrm{QART}}-s_{\mathrm{base}}}
  {s_{\mathrm{base}}}\times 100\%.
  \label{eq:relative-gain}
\end{equation}
For percentage-valued benchmarks, $\Delta$ is expressed in percentage points.
LHTB retains its native dense-reward units. Gains are calculated from unrounded
scores before display rounding.

\section{Benchmark Results}
\label{sec:results}

\subsection{Classical Baselines versus QART Hybrid Models}

Table~\ref{tab:results} compares classical baselines with QART hybrid models
across six benchmarks. Each row identifies the shared backbone LLM, while the
two score columns distinguish the classical model configuration from its QART
hybrid counterpart. This organization evaluates the QART architecture across
multiple backbone LLMs and task families.

QART hybrid models achieve higher scores in fourteen of the fifteen evaluated
backbone--benchmark pairs. The improvement is not universal: the QART model
with a DeepSeek V4 Flash backbone scores below its classical counterpart on
DeepSWE. These results provide initial evidence that the QART architecture can
improve end-to-end task performance across different backbones, while also
revealing a configuration in which performance regresses.

\begin{table*}[p]
  \centering
  \scriptsize
  \setlength{\tabcolsep}{4.5pt}
  \renewcommand{\arraystretch}{1.08}
  \caption{Performance comparison between classical baselines and QART
  quantum--classical hybrid models across six benchmarks. Each row identifies
  a shared backbone LLM. ``Classical baseline'' reports the original classical
  model configuration, while ``QART hybrid'' reports the model incorporating
  quantum encoding, CIM-based QUBO optimization, and quantum decoding layers
  within the QART architecture. All scores are our measurements in the Codex
  agent environment and may differ from model-vendor or public leaderboard
  results. Absolute change and relative gain are measured against the
  corresponding classical baseline. Percentage-valued benchmarks use
  percentage points (pp) for absolute changes; LHTB retains native reward
  units. Within each benchmark group, the highest QART score and the largest
  absolute and relative gains are shown in bold.}
  \label{tab:results}
  \begin{tabularx}{\textwidth}{
    @{}>{\raggedright\arraybackslash}p{0.15\textwidth}
    >{\raggedright\arraybackslash}p{0.20\textwidth}
    *{4}{>{\centering\arraybackslash}X}@{}}
    \toprule
    Benchmark & Backbone LLM & Classical baseline & QART hybrid & Absolute change & Relative gain \\
    \midrule
    $\tau^2$-Bench
      & GLM-5.3             & 84.11\% & 89.11\% & \textbf{+5.00 pp} & \textbf{+5.9\%} \\
      & DeepSeek V4 Flash   & 85.52\% & 88.90\% & +3.38 pp & +4.0\% \\
      & GPT-5.5 xhigh       & 88.23\% & \textbf{90.32\%} & +2.09 pp & +2.4\% \\
    \addlinespace
    $\tau^3$-Bench
      & DeepSeek V4 Flash   & 21.65\% & 31.96\% & \textbf{+10.31 pp} & \textbf{+47.6\%} \\
      & GLM-5.3             & 43.30\% & \textbf{45.36\%} & +2.06 pp & +4.8\% \\
      & GPT-5.5 xhigh       & 34.02\% & 39.18\% & +5.15 pp & +15.2\% \\
    \addlinespace
    SciCode
      & DeepSeek V4 Flash   & 17.36\% & 31.94\% & \textbf{+14.58 pp} & \textbf{+84.0\%} \\
      & GLM-5.3             & 30.90\% & \textbf{34.38\%} & +3.48 pp & +11.3\% \\
    \addlinespace
    LHTB
      & DeepSeek V4 Flash   & 0.2642 & \textbf{0.3591} & \textbf{+0.0949} & \textbf{+35.9\%} \\
      & GLM-5.3             & 0.3231 & 0.3419 & +0.0188 & +5.8\% \\
    \addlinespace
    DeepSWE
      & GPT-5.5 xhigh       & 67.00\% & \textbf{72.50\%} & +5.50 pp & +8.2\% \\
      & GLM-5.3             & 54.90\% & 65.40\% & \textbf{+10.50 pp} & \textbf{+19.1\%} \\
      & DeepSeek V4 Flash   & 45.13\% & 41.59\% & $-3.54$ pp & $-7.8$\% \\
    \addlinespace
    Terminal-Bench~4.0
      & GLM-5.3             & 14.29\% & \textbf{20.63\%} & \textbf{+6.35 pp} & \textbf{+44.4\%} \\
      & DeepSeek V4 Flash   & 14.29\% & \textbf{20.63\%} & \textbf{+6.35 pp} & \textbf{+44.4\%} \\
    \bottomrule
  \end{tabularx}
\end{table*}

\begin{figure*}[p]
  \centering
  \includegraphics[width=0.98\textwidth]{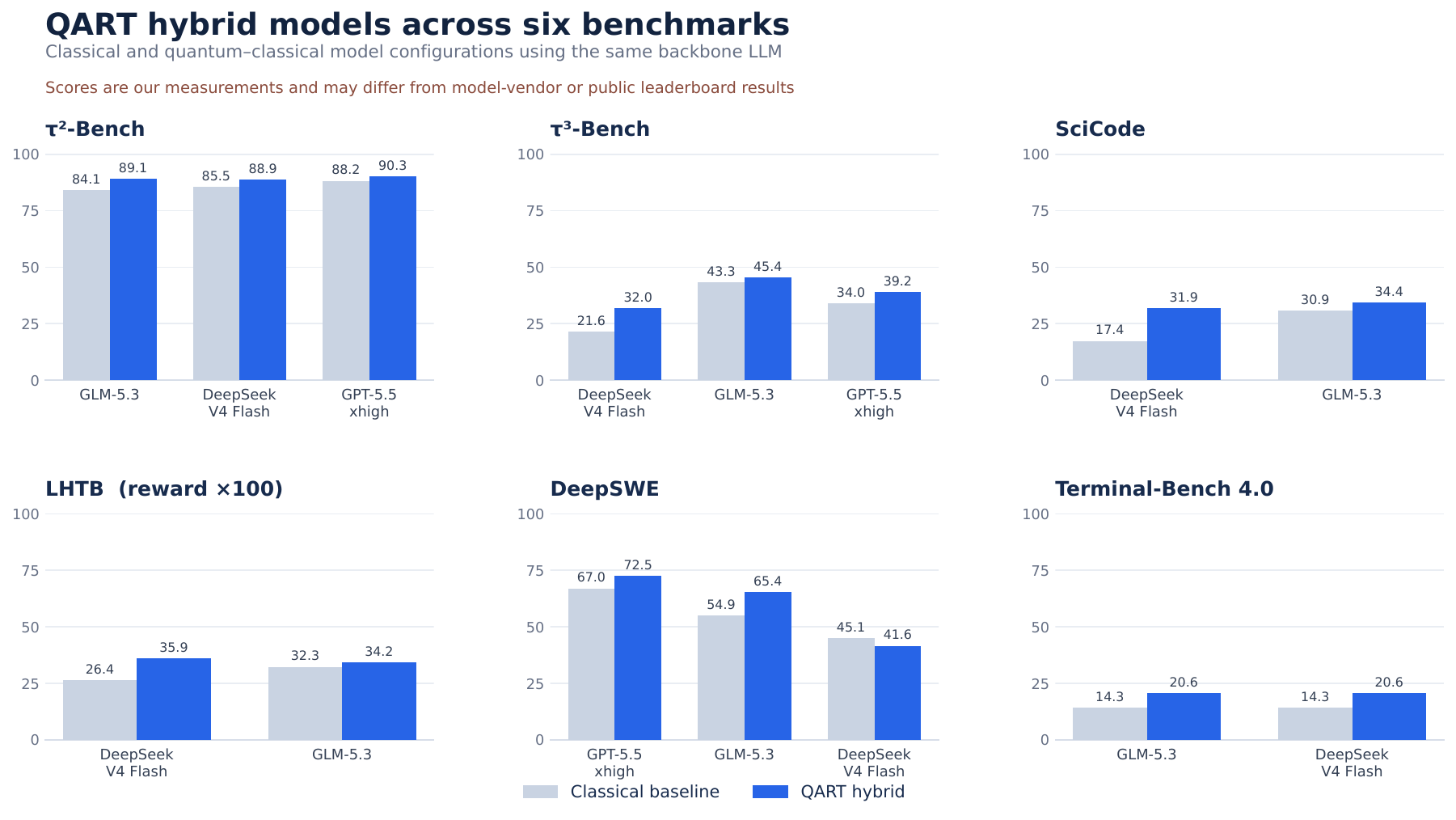}
  \caption{Classical baseline and QART hybrid model performance across six
  benchmarks. Each pair of bars compares two model configurations sharing the
  backbone LLM identified on the horizontal axis. The QART configuration
  incorporates quantum encoding, CIM-based QUBO optimization, and quantum
  decoding layers into the hybrid reasoning architecture. Scores are measured
  under the benchmark evaluation setup described in Section~\ref{sec:evaluation}.
  LHTB reward is multiplied by 100 for visualization only. All values are our
  measurements rather than model-vendor or public leaderboard scores.}
  \label{fig:performance-comparison}
\end{figure*}

Figure~\ref{fig:performance-comparison} visualizes the same paired comparisons.
LHTB reward is multiplied by 100 for display only.

\begin{figure*}[t]
  \centering
  \includegraphics[width=0.98\textwidth]{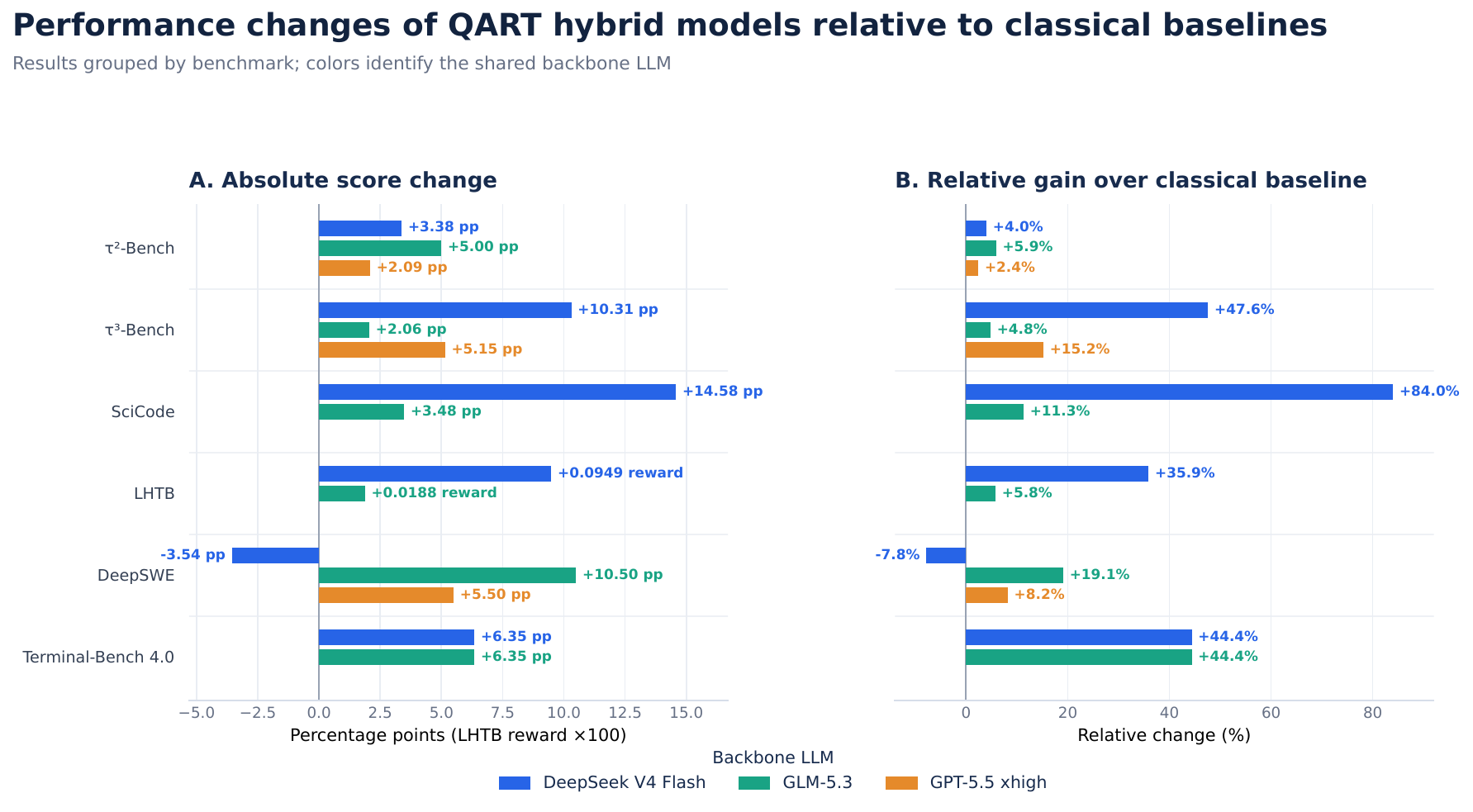}
  \caption{Performance changes from classical baselines to QART hybrid models
  constructed with the same backbone LLM. The left panel shows absolute score
  changes, and the right panel shows relative gains over the corresponding
  classical baseline. Results are grouped by benchmark, with colors identifying
  the backbone LLM. Positive values favor the QART hybrid configuration;
  negative values indicate a regression. Missing bars denote combinations that
  were not evaluated. LHTB absolute changes are multiplied by 100 for bar
  length, while annotations retain native reward units.}
  \label{fig:performance-gain}
\end{figure*}

\begin{table*}[t]
  \centering
  \small
  \setlength{\tabcolsep}{5pt}
  \renewcommand{\arraystretch}{1.10}
  \caption{Highest-scoring QART hybrid models on each benchmark and their
  closest evaluated QART counterparts. Models are identified by their
  classical backbone LLMs. All scores correspond to QART hybrid configurations
  incorporating the quantum encoding, CIM-based QUBO optimization, and quantum
  decoding layers. Parentheses report score differences from the highest-scoring
  QART model in the benchmark's native units. Rankings are restricted to the
  configurations evaluated in this report.}
  \label{tab:best-systems}
  \begin{tabularx}{\textwidth}{
    @{}>{\raggedright\arraybackslash}p{0.17\textwidth}
    >{\raggedright\arraybackslash}p{0.24\textwidth}
    >{\centering\arraybackslash}p{0.13\textwidth}
    >{\raggedright\arraybackslash}X@{}}
    \toprule
    Benchmark & Best QART backbone & Score & Closest QART model(s) \\
    \midrule
    $\tau^2$-Bench & GPT-5.5 xhigh & 90.32\% &
      GLM-5.3: 89.11\% ($-1.21$ pp); DeepSeek V4 Flash: 88.90\% ($-1.42$ pp) \\
    $\tau^3$-Bench & GLM-5.3 & 45.36\% &
      GPT-5.5 xhigh: 39.18\% ($-6.18$ pp) \\
    SciCode & GLM-5.3 & 34.38\% &
      DeepSeek V4 Flash: 31.94\% ($-2.44$ pp) \\
    LHTB & DeepSeek V4 Flash & 0.3591 &
      GLM-5.3: 0.3419 ($-0.0172$) \\
    DeepSWE & GPT-5.5 xhigh & 72.50\% &
      GLM-5.3: 65.40\% ($-7.10$ pp) \\
    Terminal-Bench~4.0 & GLM-5.3 / DeepSeek V4 Flash & 20.63\% &
      Tie: both models score 20.63\% (0.00 pp) \\
    \bottomrule
  \end{tabularx}
\end{table*}


\subsection{Benchmark-level observations}

\parahead{$\tau^2$-Bench}
The QART hybrid model with a GPT-5.5 xhigh backbone reaches the highest score,
$90.32\%$, while the GLM-5.3-backbone QART model has the largest absolute
increase, $5.00$~pp. The remaining failures are often associated with tool
arguments or communication details after the high-level operation order is
already correct.

\parahead{$\tau^3$-Bench}
The QART model with a GLM-5.3 backbone obtains the highest score at $45.36\%$.
The DeepSeek V4 Flash-backbone QART model has the largest relative gain, rising
from $21.65\%$ to $31.96\%$ ($+47.6\%$). Observed improvements include fewer
omitted steps and wrong-object actions; policy interpretation and numerical
details remain backbone-dependent.

\parahead{SciCode}
The QART model with a GLM-5.3 backbone produces the highest final score,
$34.38\%$, while the DeepSeek V4 Flash-backbone QART model shows the largest
gain in the study: $+14.58$~pp and $+84.0\%$ relative. For DeepSeek,
import-only outputs fall from 89 to 30 and syntax errors from 12 to zero.

\parahead{LHTB}
The QART hybrid model with a DeepSeek V4 Flash backbone improves from 0.2642 to
0.3591 dense reward, a $35.9\%$ relative gain averaged over three clean runs.
The GLM-5.3-backbone QART model improves from 0.3231 to 0.3419, a $5.8\%$
relative gain. Improvements are observed in multi-stage tasks such as paper
reproduction and scientific regression. Interactive tasks remain challenging
under wall-clock and tool budgets.

\parahead{DeepSWE}
The QART hybrid model with a GPT-5.5 xhigh backbone reaches the highest score in
the study, $72.5\%$, and the GLM-5.3-backbone QART model reaches $65.4\%$.
The QART model with a DeepSeek V4 Flash backbone decreases from $45.13\%$ to
$41.59\%$. Remaining failures involve implementation details such as cache
priority, parsing behavior, and serialization order. The regression shows that
the QART hybrid architecture does not eliminate every low-level implementation
error.

\parahead{Terminal-Bench~4.0}
Both the GLM-5.3-backbone and DeepSeek V4 Flash-backbone QART models improve from
9/63 tasks ($14.29\%$) to 13/63 ($20.63\%$), corresponding to a $44.4\%$
relative gain. More terminal tasks are completed successfully, while geometry,
numerical, and fine-grained rule errors still dominate the shared failures.

\subsection{Absolute and relative gains}

Figure~\ref{fig:performance-gain} groups the two gain views by benchmark on the
vertical axis. Within each benchmark group, distinct bar colors identify the
shared backbone LLM, so the comparison remains within the same task and
verifier. SciCode illustrates why both metrics are needed: a $14.58$-point
increase becomes an $84.0\%$ relative gain because the classical baseline is
low. Across the reported pairs, the median relative change is $11.3\%$, but the
negative DeepSWE case shows that an aggregate summary must not be interpreted as
a per-task guarantee.

\subsection{Comparison across QART Backbone Instantiations}

Table~\ref{tab:best-systems} compares QART hybrid models instantiated with
different classical backbones. The highest-scoring backbone varies across
benchmarks, indicating that the resulting hybrid model's performance remains
dependent on both the backbone LLM and the task family. These comparisons
characterize the evaluated QART configurations and do not establish a ranking
against unevaluated systems.

These comparisons evaluate the complete QART hybrid model end to end; the
independent contribution of the quantum optimization layers would require a
classical-solver control under the same encoding to identify.

\FloatBarrier
\section{Limitations}
\label{sec:limitations}

The paired results characterize end-to-end performance of classical baselines
and QART hybrid models built with the same backbone LLM. The report describes the system's functional
organization and evaluation protocol, while its internal information
processing and optimization implementation remain proprietary. This limits
independent reproduction of the internal method and attribution of observed
changes to individual mechanisms.


Scores measured under our Codex agent and harness should not be treated as
interchangeable with vendor or public leaderboard scores. Changes in the agent,
tools, prompts, background context, retry budget, or verifier version can alter
the absolute level even when the model name is unchanged. The DeepSWE
regression further shows that the QART hybrid configuration does not
consistently improve every evaluated setting or eliminate low-level
implementation errors.


\section{Conclusion}
\label{sec:conclusion}

\QART, the Quantum-Augmented Reasoning Transformer, is a quantum--classical
hybrid architecture that combines a backbone language model with an
optimization component. The architecture assigns task understanding,
reasoning, and generation to the language model and uses quantum encoding,
CIM-based QUBO optimization, and quantum decoding as functional layers for
supplying auxiliary information to subsequent model activity.
The report focuses on these observable functional roles; the detailed
information-processing and optimization implementations remain proprietary.

The resulting conditional theorem gives QART a non-vanishing path-recovery
probability and a positive asymptotic reliability gap over single-trajectory
autoregressive reasoning.

Across six long-horizon benchmarks, fourteen of fifteen evaluated
backbone--benchmark pairs favor the QART hybrid model over the corresponding
classical baseline. The largest relative gains are $84.0\%$ on SciCode,
$47.6\%$ on $\tau^3$-Bench, and $44.4\%$ on Terminal-Bench~4.0. The QART hybrid
model with a GPT-5.5 xhigh backbone reaches $90.32\%$ on $\tau^2$-Bench and
$72.5\%$ on DeepSWE, while the GLM-5.3-backbone QART model provides the best
measured $\tau^3$-Bench and SciCode scores. The DeepSeek V4 Flash-backbone QART
model decreases by $-7.8\%$ on DeepSWE, demonstrating that the observed effect
varies across configurations and does not eliminate backbone-specific
implementation failures.

These measurements provide initial system-level evidence for the evaluated QART
configurations and motivate further study of the relationship between
solver-specific effective optimization capacity and externally measured
reasoning horizons. The conditional scaling framework in this report provides a
basis for organizing that work, including matched solver comparisons and
resource-scale experiments that account for the full end-to-end evaluation
cost.

\section*{Acknowledgements}
The authors used ChatGPT to assist with drafting and editing the manuscript.
ChatGPT was also used as an aid in developing and checking the proofs in
Section 5 and Appendix A. The authors reviewed, verified, and take full
responsibility for all scientific claims, mathematical arguments, experimental
results, and the final text.

\renewcommand{\bibfont}{\footnotesize}
\bibliographystyle{unsrtnat}
\bibliography{references}

\appendix
\section*{Appendix}
\addcontentsline{toc}{section}{Appendix}
\section{Proofs and Technical Conditions}
\label{app:theory-proofs}

This appendix establishes the statements used in Section~\ref{sec:theory} and isolates the assumptions needed to extend them. The spectral certificate is a reformulation of the result attributed to Wang et al.~\citep{wang2023bifurcation}. The semantic transfer and probability composition are derived here from the stated interface properties. All guarantees concern declared task families and resource schedules.

\subsection{Mathematical Setting and Notation}
\label{app:setting}

For each horizon, take a probability space supporting the task, semantic representation, encoded instance, and physical run. The path space and objective are fixed by the task specification before an encoded instance is evaluated. They may vary with the sampled task, although this dependence is suppressed in the notation. Assume all events and maps used below are measurable. Ground-state and path minima exist because their declared search spaces are finite and nonempty.

When a conditioning event has probability zero, the corresponding factor need not be defined; the certified intersection already has probability zero and the nontrivial bound is unavailable. For the positive uniform result all relevant prefixes of the event intersection have positive probability. Similarly, zero-probability survival in Proposition~1 directly implies zero final path probability.

The notation table at the end of this appendix records dimensions and separates task horizon, token count, and encoded problem size. The scalar $p_{0,h}$ is a bifurcation threshold, whereas $p_\star$ is a probability lower bound. Neither should be confused with an empirical solver success target.

\subsection{Proof of Proposition 1}
\label{app:prop-proof}

Because continuation extends the committed history, a prefix that can be completed to an acceptable trajectory must have had a completable predecessor. Therefore $C_{h,j}\subseteq C_{h,j-1}$. If every predecessor has positive probability, repeated conditioning gives
\begin{equation}
 \Pr(C_{h,m(h)})=\prod_{j=1}^{m(h)}\Pr(C_{h,j}\mid C_{h,j-1}).
 \label{eq:app-ar-product}
\end{equation}
Substitution of Eq.~\eqref{eq:theory-risk} proves Eq.~\eqref{eq:ar-product}. If any factor is zero, final success is zero. Otherwise, taking logarithms and using $\log(1-x)\le-x$ gives
\begin{equation}
 \log P_{\rm AR}^{\rm path}(h)\le-\sum_{j=1}^{m(h)}e_{h,j}=-E_h.
 \label{eq:app-ar-log}
\end{equation}
Exponentiation yields Eq.~\eqref{eq:ar-bound}. Since $E_h\to\infty$, its upper bound tends to zero. The probability is nonnegative, so it also tends to zero. This proves Proposition~1.

The assumption concerns the sum of conditional risks, not merely the existence of occasional errors. If each of $h$ decisions has risk $h^{-2}$, then $E_h=h^{-1}$ and the survival product tends to one. If each risk is $h^{-1}$, the product tends to $e^{-1}$. These examples show why increasing depth alone is insufficient for a vanishing-success theorem. They also illustrate that the condition should be assessed on the policy and resource schedule actually compared.

\subsection{Spectral Proof of Theorem NC and Its Dynamical Meaning}
\label{app:spectral-proof}

Suppress the horizon index and write $n=n_h$, $G=G_h$, $\mathbf v=\mathbf v_h$, and $\boldsymbol\sigma=\operatorname{sign}(\mathbf v)$. The nonzero-coordinate assumption ensures $\|\boldsymbol\sigma\|_2^2=n$. Let $\mathbf u=\boldsymbol\sigma/\sqrt n$. Decompose the unit vector $\mathbf u$ into the top eigendirection and its orthogonal complement:
\begin{equation}
 \mathbf u=c\mathbf v+\mathbf z,\qquad \mathbf v^{\mathsf T}\mathbf z=0,
 \label{eq:app-decomp}
\end{equation}
where $c=\mathbf v^{\mathsf T}\mathbf u$ and $\|\mathbf z\|_2^2=1-c^2$. By Eq.~\eqref{eq:nc-overlap}, $c^2=\alpha_h^2$. Orthogonality eliminates cross terms and the Rayleigh bound gives
\begin{equation}
 \begin{aligned}
 \mathbf u^{\mathsf T}G\mathbf u
 &\ge\lambda_{\max}c^2+\lambda_{\min}(1-c^2)\\
 &=\lambda_{\max}\alpha_h^2+\lambda_{\min}(1-\alpha_h^2).
 \end{aligned}
 \label{eq:app-rayleigh}
\end{equation}
Consequently, with $w=\lambda_{\max}-\lambda_{\min}$,
\begin{equation}
 H(\boldsymbol\sigma)\le-\frac n2\lambda_{\max}+\frac{nw}{2}(1-\alpha_h^2).
 \label{eq:app-certified-energy}
\end{equation}
For every binary vector $\boldsymbol\tau$, the upper Rayleigh bound implies $H(\boldsymbol\tau)\ge-n\lambda_{\max}/2$. In particular,
\begin{equation}
 H_1=H_0+\Delta H\ge-\frac n2\lambda_{\max}+\Delta H.
 \label{eq:app-excited-bound}
\end{equation}
The strict condition in Eq.~\eqref{eq:nc-certificate} makes the right-hand side of Eq.~\eqref{eq:app-certified-energy} smaller than the right-hand side of Eq.~\eqref{eq:app-excited-bound}. Hence $H(\boldsymbol\sigma)<H_1$. By definition, no non-ground-state binary configuration has energy below $H_1$, and thus $H(\boldsymbol\sigma)=H_0$. The Hamiltonian is invariant under global sign reversal, so $-\boldsymbol\sigma$ is also a ground state.

For the deterministic dynamics, the Jacobian at the origin is
\begin{equation}
 J(0,p)=(p-1)I+\xi G.
 \label{eq:app-jacobian}
\end{equation}
Its largest eigenvalue crosses zero at $p_0=1-\xi\lambda_{\max}$. If the maximum eigenvalue is simple, the remaining linear modes are stable at this threshold. Projecting a nearby equilibrium onto $\mathbf v$ gives the leading amplitude equation. Writing the positive branch amplitude as $r>0$, the two local branches satisfy
\begin{equation}
 \mathbf x_\pm(p)=\pm r(p)\mathbf v+o(r(p)),\qquad r(p)>0,
 \label{eq:app-branches}
\end{equation}
and
\begin{equation}
 0=(p-p_0)r-r^3\sum_i v_i^4+o(r^3).
 \label{eq:app-amplitude}
\end{equation}
Thus the small nonzero branches have leading direction $\pm\mathbf v$, with amplitude proportional to $\sqrt{p-p_0}$. More precisely, $\mathbf x_\pm(p)/r(p)\to\pm\mathbf v$ as $p\downarrow p_0$. This limiting statement is sufficient: if $\mu=\min_i|v_i|>0$, a sufficiently small directional perturbation preserves all signs.

The argument does not say that a finite-amplitude equilibrium is exactly a scalar multiple of $\mathbf v$, that an arbitrary initialization reaches that branch, or that a finite-rate pump follows it. Nor does it transfer a certificate automatically to a different feedback model or control schedule. Those implementation questions are assigned to the reachability and readout events. Existence of a stable ground-state equilibrium elsewhere in parameter space would also be insufficient by itself to establish its sampling probability.

\subsection{Proof of Lemma 1 and an Approximate-Solution Extension}
\label{app:semantic-proof}

Let $\boldsymbol\sigma^\star$ be a valid encoding of an optimal path, whose existence is required by Assumption~A. For any valid $\boldsymbol\tau$ decoding to a non-optimal path, semantic fidelity gives
\begin{equation}
 H_h(\boldsymbol\tau)\ge a_h(L_h^\star+\gamma_h)+b_h-\epsilon_h,
 \label{eq:app-nonoptimal-lower}
\end{equation}
\begin{equation}
 H_h(\boldsymbol\sigma^\star)\le a_hL_h^\star+b_h+\epsilon_h.
 \label{eq:app-optimal-upper}
\end{equation}
Subtracting yields
\begin{equation}
 H_h(\boldsymbol\tau)-H_h(\boldsymbol\sigma^\star)\ge a_h\gamma_h-2\epsilon_h>0.
 \label{eq:app-semantic-separation}
\end{equation}
Hence no non-optimal valid state minimizes the energy. By assumption, no invalid state minimizes it either. Every ground state therefore decodes to an optimal path. This proves Lemma~1.

The result does not require equal encoded energies among all task-optimal paths. Their energies can differ within the fidelity tolerance; some optimal paths may therefore fail to be ground states. The necessary implication is from every ground state to an optimal path, not the reverse.

An approximate energy guarantee also transfers. Suppose $\widehat{\boldsymbol\sigma}$ is valid and satisfies
\begin{equation}
 H_h(\widehat{\boldsymbol\sigma})\le H_{0,h}+\delta_h,\qquad \delta_h\ge0.
 \label{eq:app-approx-energy}
\end{equation}
Since $H_{0,h}\le H_h(\boldsymbol\sigma^\star)$, applying fidelity to both states gives
\begin{equation}
 L_h(D_h(\widehat{\boldsymbol\sigma}))-L_h^\star\le\frac{\delta_h+2\epsilon_h}{a_h}.
 \label{eq:app-approx-objective}
\end{equation}
If $\delta_h+2\epsilon_h<a_h\gamma_h$, the returned valid state must still decode to an optimal path. Validity must be checked or guaranteed separately: an energy tolerance can admit an invalid state even when exact ground states are valid. This extension connects an energy-tolerance notion of effective optimization capacity with semantic performance without changing the exact-recovery theorem.

\subsection{Proof of Theorem 1 and a Precise Reachability Event}
\label{app:theorem-proof}

On $\mathsf M_h\cap\mathsf S_h$, the two certified sign patterns are ground states, and both decode to optimal paths. Put $\mu_h=\min_i|v_{h,i}|$. One sufficient realization of $\mathsf B_h$ is that, by the prescribed measurement time, the physical state $\mathbf x$ satisfies, for some scale $a>0$ and sign $s\in\{-1,+1\}$,
\begin{equation}
 \left\|\frac{\mathbf x}{a}-s\mathbf v_h\right\|_\infty<\frac{\mu_h}{2},
 \label{eq:app-sign-neighborhood}
\end{equation}
and retains this sign pattern until readout. Then every coordinate has sign $s\operatorname{sign}(v_{h,i})$. The scale $a$ is an amplitude normalization, not the semantic scaling coefficient $a_h$. Equivalent implementation-specific sufficient events can be used if they ensure the same certified signs within budget.

Define the certified success intersection
\begin{equation}
 \mathsf F_h=\mathsf M_h\cap\mathsf S_h\cap\mathsf B_h\cap\mathsf R_h.
 \label{eq:app-certified-intersection}
\end{equation}
On this event, Theorem NC, faithful readout, and Lemma~1 imply that the decoded path lies in $\Pi_h^\star$. Therefore
\begin{equation}
 P_{\rm QART}^{\rm opt}(h)\ge\Pr(\mathsf F_h).
 \label{eq:app-opt-lower}
\end{equation}
Successive conditional factorization gives
\begin{equation}
 \Pr(\mathsf F_h)=r_M(h)r_S(h)r_B(h)r_R(h).
 \label{eq:app-factorization}
\end{equation}
This proves the finite-horizon bound. If the four factors have the stated uniform positive lower bounds for every $h\ge h_0$, then $P_{\rm QART}^{\rm opt}(h)\ge p_\star$ for every such $h$. Taking a limiting inferior proves Eq.~\eqref{eq:uniform-bound}.

Uniformity is the demanding step. It can fail even when the solver is perfect on every represented instance: for example, if coverage decays to zero, the certified intersection may also vanish. Similarly, an ideal deterministic spectral certificate cannot establish a nonzero uniform physical probability without information about initialization, noise, time, and precision. The factorization is exact for the certified event; the substantive content of an application lies in proving or supporting the factors.

\subsection{Reliability Separation and End-to-End Correctness}
\label{app:separation}

Assume objective alignment $\Pi_h^\star\subseteq\mathcal A_h$ and the common evaluation setting described in Section~\ref{sec:reliability-separation}. The event of optimal recovery implies acceptable path recovery. Thus, for sufficiently large $h$,
\begin{equation}
 P_{\rm QART}^{\rm path}(h)\ge P_{\rm QART}^{\rm opt}(h)\ge p_\star.
 \label{eq:app-path-bound}
\end{equation}
Proposition~1 gives $P_{\rm AR}^{\rm path}(h)\to0$. For any $\varepsilon>0$, this probability is below $\varepsilon$ eventually; consequently the difference in path-success probabilities is eventually at least $p_\star-\varepsilon$. Letting $\varepsilon\downarrow0$ proves
\begin{equation}
 \liminf_{h\to\infty}\left[P_{\rm QART}^{\rm path}(h)-P_{\rm AR}^{\rm path}(h)\right]\ge p_\star.
 \label{eq:app-path-separation}
\end{equation}
For final-answer correctness, require $\Pr(\mathsf X_h\mid\mathsf F_h)\ge\rho_X>0$ uniformly. For all sufficiently large $h$, the preceding bound on $\Pr(\mathsf F_h)$ then gives
\begin{equation}
 \Pr(\mathsf X_h)\ge\Pr(\mathsf X_h\cap\mathsf F_h)\ge\rho_Xp_\star.
 \label{eq:app-answer-bound}
\end{equation}
This does not require independence between selection and execution. Without the conditional execution bound, path recovery does not entail a non-vanishing final-answer probability. Conversely, a baseline can sometimes produce a correct final answer despite an unacceptable internal trajectory, so an exact-path decay theorem cannot automatically be relabeled as a final-answer decay theorem.

\subsection{Perturbation Bounds and the Limits of Local Robustness}
\label{app:perturbation}

For any binary vector, $\|\boldsymbol\sigma\|_2^2=n_h$. The operator-norm inequality therefore gives
\begin{equation}
 \frac12\left|\boldsymbol\sigma^{\mathsf T}\Delta G_h\boldsymbol\sigma\right|\le\frac{n_h}{2}\eta_h,
 \label{eq:app-perturbation-energy}
\end{equation}
which proves Eq.~\eqref{eq:perturbation-energy}. For an original ground state $\boldsymbol\sigma_0$ and any original non-ground state $\boldsymbol\tau$,
\begin{equation}
 H_{\widetilde G_h}(\boldsymbol\tau)-H_{\widetilde G_h}(\boldsymbol\sigma_0)\ge\Delta H_h-n_h\eta_h.
 \label{eq:app-perturbed-gap}
\end{equation}
Thus $n_h\eta_h<\Delta H_h$ prevents any original non-ground state from becoming a perturbed minimizer. Original ground states can split in energy, so the conclusion is inclusion of the perturbed ground-state set in the original set. If original ground states decode optimally, all perturbed minima continue to decode optimally under the same decoder.

A direct semantic bound can be less restrictive. Let $f_h>0$ be the energy gap from the minimum valid state to the lowest invalid state, with $f_h=+\infty$ if no invalid states exist. With the decoder fixed, it is sufficient that
\begin{equation}
 n_h\eta_h<\min\{a_h\gamma_h-2\epsilon_h,\ f_h\}.
 \label{eq:app-semantic-perturbation}
\end{equation}
The first term keeps every non-optimal valid state above an optimal encoding; the second keeps invalid states above a valid minimum. This guarantees task-optimal perturbed ground states without requiring all original energy levels to retain their ordering.

For stability of the certified eigenvector signs, let $\kappa_h=\lambda_{\max,h}-\lambda_{2,h}>0$ and assume $\eta_h<\kappa_h/2$. The largest perturbed eigenvalue remains simple by the variational eigenvalue bound. Align its unit eigenvector $\widetilde{\mathbf v}_h$ with $\mathbf v_h$. Projecting the perturbed eigenvalue equation onto the orthogonal complement of $\mathbf v_h$ yields
\begin{equation}
 \sin\angle(\widetilde{\mathbf v}_h,\mathbf v_h)\le\frac{\eta_h}{\kappa_h-\eta_h}.
 \label{eq:app-eigen-angle}
\end{equation}
Indeed, on that complement the unperturbed matrix has eigenvalues at most $\lambda_{2,h}$, whereas the perturbed leading eigenvalue is at least $\lambda_{\max,h}-\eta_h$. Inverting this restricted operator gives Eq.~\eqref{eq:app-eigen-angle}. For aligned unit vectors,
\begin{equation}
 \|\widetilde{\mathbf v}_h-\mathbf v_h\|_2\le\frac{2\sqrt2\,\eta_h}{\kappa_h}.
 \label{eq:app-eigenvector-distance}
\end{equation}
If the right-hand side is below $\mu_h$, all eigenvector signs agree. If, in addition, the original ground-state set consists only of the certified pair $\{\pm\boldsymbol\sigma_h^c\}$ and $n_h\eta_h<\Delta H_h$, that pair remains the perturbed ground-state set. With additional degenerate ground states, sign preservation alone does not prove that the same pair minimizes the perturbed energy.

These qualifications also matter for the NC inequality itself. Splitting a degenerate ground-state level can create a much smaller perturbed gap, so preservation of the original certificate threshold does not follow merely from small matrix error. One must verify the perturbed certificate or establish an alternative semantic and dynamical certificate. Finally, every radius above may shrink with $h$; a finite-instance neighborhood is not a dimension-independent noise tolerance.

\subsection{Resource Schedules and Repeated Sampling}
\label{app:resources}

Let all four conditional factors depend on a resource schedule $B(h)$ and on physical precision, while suppressing those arguments for brevity. A uniform theorem can follow if the task family, encoder, hardware scaling, and control schedule jointly keep the certified intersection probability bounded away from zero. The theorem does not prescribe how large $B(h)$ must be, and it makes no polynomial-time claim.

Repeated solver sampling can improve recovery on a fixed encoded instance. If every attempt has conditional ground-state success probability at least $q_h^{\rm hit}$ even given all previous failures, then the probability that at least one of $K$ attempts succeeds is at least
\begin{equation}
 1-(1-q_h^{\rm hit})^K.
 \label{eq:repeated-sampling}
\end{equation}
This follows by factoring the probability of repeated failure; independence is sufficient but not necessary. Choosing the lowest-energy valid sample preserves a ground-state hit when energies and validity are evaluated faithfully. No knowledge of the exact ground-state energy is needed for this selection rule, though certification of the final result is a separate issue.

Repeated physical attempts do not repair an encoding that omits all true optima or reverses their semantic ranking. Regenerating semantic representations is a different intervention whose coverage probability and cost must also be measured. Furthermore, if $q_h^{\rm hit}$ decreases rapidly, the number of attempts needed to maintain a target success probability can grow rapidly and invalidate a fixed-budget claim.

For readout, a simple illustration is independent sign errors of probability $\nu_h$ across $n_h$ measured coordinates. The probability of no sign error is $(1-\nu_h)^{n_h}$, which tends to zero as $n_h\to\infty$ when $\nu_h\equiv\nu\in(0,1)$ is fixed and positive. Redundancy, error detection, amplitude control, or a decreasing error rate may change that behavior, but require separate evidence. Neither a high finite-size success rate nor a large advertised oscillator count establishes the uniform readout factor.

\subsection{Notation and Dimensions}
\label{app:notation}

The dimensions below apply to each fixed task and encoding realization. Token count, representation size, and spin count may all depend on the horizon.

\begin{center}
\scriptsize
\setlength{\tabcolsep}{2pt}
\tablefirsthead{\toprule Symbol & Meaning and dimension \\ \midrule}
\tablehead{\toprule Symbol & Meaning and dimension \\ \midrule}
\tabletail{\midrule}
\tablelasttail{\bottomrule}
\begin{supertabular}{@{}p{0.36\columnwidth}p{0.60\columnwidth}@{}}
$h,\ m(h)$ & Reasoning horizon and number of critical decisions; positive integers. \\
$q_h,\ y_h$ & Task specification and generated textual information; elements of task and text spaces. \\
$\mathcal A_h,\ \Pi_h,\ \Pi_h^\star$ & Acceptable trajectories, declared admissible paths, and objective minimizers; finite sets in this formulation. \\
$C_{h,j},\ e_{h,j},\ E_h$ & Prefix-survival event, scalar conditional exit probability, and scalar cumulative risk. \\
$Z_h$ & Accessible hidden states; for a single layer, $Z_h\in\mathbb R^{\ell_h\times d}$, with token count $\ell_h$ and hidden width $d$. \\
$u_h,\ \mathcal U_h$ & Semantic information and its representation space; if vector-valued, $u_h\in\mathbb R^{d_u(h)}$. \\
$\Phi_{\rm hid},\ \Phi_{\rm text}$ & Abstract semantic extraction maps with hidden-state or text input. \\
$\mathcal E,\ D_h$ & Abstract encoder to a symmetric zero-diagonal matrix and decoder to a path or invalid symbol. \\
$n_h,\ G_h$ & Number of encoded Ising variables, including auxiliaries; $G_h\in\mathbb R^{n_h\times n_h}$. \\
$\boldsymbol\sigma,\ \boldsymbol\sigma_h^c$ & Binary spin vector and certified sign pattern; both in $\{-1,+1\}^{n_h}$. \\
$\boldsymbol\sigma^\star,\ \widehat{\boldsymbol\sigma}$ & Valid encoding of a task-optimal path and returned valid approximate solution; both in $\{-1,+1\}^{n_h}$. \\
$\mathbf x,\ \mathbf v_h$ & Physical amplitude vector and unit leading eigenvector; both in $\mathbb R^{n_h}$. \\
$\widetilde G_h,\ \widetilde{\mathbf v}_h$ & Perturbed coupling matrix and its aligned unit leading eigenvector; dimensions $n_h\times n_h$ and $n_h$. \\
$H_h,\ H_{0,h},\ H_{1,h},\ \Delta H_h$ & Scalar Ising energy function, ground energy, first distinct excited energy, and positive energy gap. \\
$H_{G_h},\ H_{\widetilde G_h}$ & Energy functions associated with $G_h$ and $\widetilde G_h$; $H_{G_h}\equiv H_h$. \\
$L_h,\ L_h^\star,\ \gamma_h$ & Scalar task objective, its minimum, and gap to the best non-optimal path. \\
$a_h,\ b_h,\ \epsilon_h$ & Positive semantic energy scale, scalar energy offset, and nonnegative fidelity error. \\
$p,\ p_{0,h},\ \xi$ & Scalar pump parameter, first bifurcation threshold, and positive coupling scale. \\
$\lambda_{\max,h},\ \lambda_{2,h},\ \lambda_{\min,h}$ & Largest, second largest, and smallest eigenvalues of $G_h$; real scalars. \\
$w_h,\ \kappa_h,\ \mu_h,\ c_h$ & Spectral width, leading spectral gap, minimum eigenvector magnitude, and certificate slack; scalars. \\
$\alpha_h^2$ & Scalar synchronization statistic in $[0,1]$ for the normalized leading mode. \\
$\mathsf M_h,\mathsf S_h,\mathsf B_h,\mathsf R_h$ & Fidelity, spectral certification, arrival, and faithful readout events. \\
$\nu_h$ & Per-coordinate readout sign-error probability; dimensionless scalar. \\
$r_i(h),\ \rho_i,\ p_\star$ & Conditional probabilities, uniform positive lower bounds, and their product; dimensionless scalars. \\
$\mathsf F_h,\mathsf X_h,\ \rho_X$ & Certified intersection, correct final execution event, and its conditional lower bound. \\
$\Delta G_h,\ \eta_h$ & Matrix perturbation in $\mathbb R^{n_h\times n_h}$ and its scalar operator norm. \\
$\delta_h,\ f_h$ & Allowed energy suboptimality and invalid-state energy margin; scalars. \\
$B(h),\ K,\ q_h^{\rm hit}$ & Resource budget, number of solver attempts, and per-attempt hit probability. \\
\end{supertabular}
\end{center}

\end{document}